\documentclass[conference]{IEEEtran}
\pdfoutput=1    

\makeatletter
\def\ps@IEEEtitlepagestyle{%
\def\@oddfoot{\mycopyrightnotice}%
\def\@evenfoot{}%
}
\def\mycopyrightnotice{%
{\footnotesize 978-1-6654-7095-7/22/\$31.00~\copyright~2022 IEEE\hfill}
\gdef\mycopyrightnotice{}
}

\usepackage{blindtext}
\usepackage{eso-pic}
\usepackage{svg}

\usepackage[ruled,linesnumbered]{algorithm2e}

\IEEEoverridecommandlockouts

\usepackage{xcolor}
\definecolor{wong-black}        {HTML}{000000}
\definecolor{wong-lightorange}  {HTML}{E69F00}
\definecolor{wong-lightblue}    {HTML}{56B4E9}
\definecolor{wong-green}        {HTML}{009E73}
\definecolor{wong-yellow}       {HTML}{F0E442}
\definecolor{wong-darkblue}     {HTML}{0072B2}
\definecolor{wong-darkorange}   {HTML}{D55E00}
\definecolor{wong-pink}         {HTML}{CC79A7}

\usepackage{url}

\usepackage{hyperref} 
\hypersetup{
    colorlinks=true,
    citecolor=wong-green,
    linkcolor=wong-darkblue,
    filecolor=wong-pink,      
    urlcolor=wong-black,
    pdfpagemode=FullScreen,
    }

\usepackage{cite}
\usepackage{amsmath,amssymb,amsfonts}
\usepackage{algorithmic}
\usepackage{graphicx}
\usepackage{textcomp}
\usepackage[nolist, nohyperlinks, printonlyused]{acronym} 
\usepackage{listings}

\def\BibTeX{{\rm B\kern-.05em{\sc i\kern-.025em b}\kern-.08em
    T\kern-.1667em\lower.7ex\hbox{E}\kern-.125emX}}

\usepackage{eso-pic}
\newcommand\AtPageUpperMyright[1]{\AtPageUpperLeft{%
 \put(\LenToUnit{0.17\paperwidth},\LenToUnit{-2cm}){%
     \parbox{0.9\textwidth}{\raggedleft\fontsize{8}{11}\selectfont #1}}%
 }}%
\newcommand{\conf}[1]{%
\AddToShipoutPictureBG*{%
\AtPageUpperMyright{#1}
}
}

\begin{document}


\title{\vspace*{1cm} DLCSS: Dynamic Longest Common Subsequences}


\author{\IEEEauthorblockN{Daniel Bogdoll\IEEEauthorrefmark{2}\IEEEauthorrefmark{3},
Jonas Rauch\IEEEauthorrefmark{3},
and J. Marius Zöllner\IEEEauthorrefmark{2}\IEEEauthorrefmark{3}}

\IEEEauthorblockA{\IEEEauthorrefmark{2}FZI Research Center for Information Technology, Germany\\
bogdoll@fzi.de}
\IEEEauthorblockA{\IEEEauthorrefmark{3}Karlsruhe Institute of Technology, Germany\\}}



\maketitle
\conf{\textit{Proc. of the International Conference on Electrical, Computer, Communications and Mechatronics Engineering  (ICECCME) \\ 
16-18 November 2022, Maldives}}


\begin{acronym}
    \acro{ml}[ML]{Machine Learning}
	\acro{cnn}[CNN]{Convolutional Neural Network}
	\acro{dl}[DL]{Deep Learning}
	\acro{ad}[AD]{Autonomous Driving}
\end{acronym}


\begin{abstract}
Autonomous driving is a key technology towards a brighter, more sustainable future. To enable such a future, it is necessary to utilize autonomous vehicles in shared mobility models. However, to evaluate, whether two or more route requests have the potential for a shared ride, is a compute-intensive task, if done by rerouting. In this work, we propose the Dynamic Longest Common Subsequences algorithm for fast and cost-efficient comparison of two routes for their compatibility, dynamically only incorporating parts of the routes which are suited for a shared trip. Based on this, one can also estimate, how many autonomous vehicles might be necessary to fulfill the local mobility demands. This can help providers to estimate the necessary fleet sizes, policymakers to better understand mobility patterns and cities to scale necessary infrastructure. 

\end{abstract}




\section{Introduction}
\label{sec:introduction}

Autonomous vehicles in the field of public transport will most likely operate in shared mobility models in the future \cite{Uber, Lyft}. This enables a sustainable usage of this groundbreaking technology. Therefore, it is necessary to match the desired routes of two or more passengers for compatibility. While solutions exist to determine the compatibility between two routes by computing the exact detour that occurs when a second route is considered \cite{googlemapssdk, mapboxsdk}, this represents a multistep, computationally intensive procedure that requires recomputation of routes with a routing engine. Given multiple routes, it is desirable to determine a similarity score to determine whether any two routes are suitable for a joint trip. For this purpose, in this work we propose the Dynamic Longest Common Subsequences (DLCSS) algorithm to dynamically find the longest common subsequences of two routes and based on this, determine a compatibility for a common trip with a similarity metric $sm_{DLCSS}$. This work is also a step towards determining realistic estimates for traffic volumes in a future with autonomous vehicles. This can help guide the progress of autonomous vehicles, e.g., by using such estimates to properly size necessary network infrastructure~\cite{Bogdoll_KIGLIS_2021_ISC2,bundesregierungEntwurfGesetzesZur2021}. The presented concept was already developed in~\cite{bogdoll2019MA} as an unpublished thesis of the first author. With this publication, we extend the state-of-the-art research with current literature, provide the DLCSS algorithm in pseudocode and aim to make the concept available to a broad, international audience.

\subsection{Contribution}
With this work, on a high level, we aim to contribute to the Sustainable Development Goals (SDG)~\cite{sdg}. The vision of tomorrow's shared, autonomous mobility can contribute to numerous SDGs, and our work is a building block towards that vision. Based on the analyses provided by \cite{didisdg, milessdg}, the SDGs 1, 5, and 8 – 13 can be supported by shared mobility providers: No poverty; Gender Equality; Decent Work and Economic Growth; Industry, Innovation and Infrastructure; Reduced Inequalities; Sustainable Cities and Communities; Responsible Consumption and Production; and Climate Action. On a more technical level, we contribute to the general field of route and trajectory matching, which can also find application outside the field of autonomous driving. While existing methods often lack temporal components of trajectories or work in a static, parameter-based fashion, our proposed algorithm is a dynamic, parameter-free approach to evaluate two routes for their suitability of a shared trip.\\
In Section~\ref{sec:related_work}, we provide an overview of route comparison approaches. Section~\ref{sec:method} describes our proposed algorithm. Here, we go into detail with a pseudocode implementation and visualize the underlying principle. In Section~\ref{sec:evaluation}, we perform a series of experiments and evaluate the algorithm. Finally, Section~\ref{sec:conclusion} provides a summary and outlook.

\begin{figure}[t!]
\centering
\resizebox{\linewidth}{!}{
\includegraphics{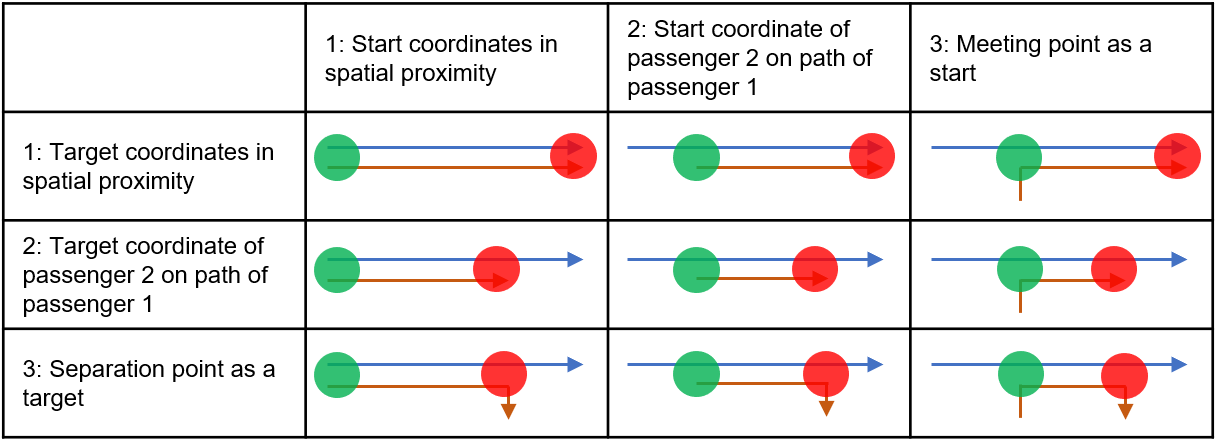}}
\caption{Scenario overview of how two people can share a ride. In [1,1] they share the whole route. In [1,2], the start coordinate of the second passenger is on the route of the first. In [1,3], the second passenger gets to a meeting point on or near the route of the first passenger by other means of transport or by foot. In [2,1], the destination of the second rider is on the route of the first rider. In [2,2], the second person's route is a partial route of the first. In [2,3], the second person comes to a meeting point and has their destination on the route of the first. [3,1] shows a common starting point, while the second person reaches their destination differently from a separation point. [3,2] is similar, but the starting point of rider 2 is on the route of rider 1. In the most complex scenario [3,3], rider 2 joins via a meeting point and continues their journey after a separation point. Adapted from \cite{bogdoll2019MA}.}
\label{fig:scenarios}
\end{figure}

\section{Related Work}
\label{sec:related_work}


For the comparison of two routes, numerous metrics exist in the literature. For this survey, we focus on sophisticated methods for route similarity measures. These either evaluate spatial-temporal similarity or only spatial similarity. In \cite{Magdy2017AGT}, a detailed analysis of twelve published metrics was performed. To evaluate routes based on their partial overlap, the Fréchet metric \cite{Toohey2015}, dynamic time warping (DTW), edit distance, and longest common subsequence (LCSS) methods \cite{Vlachos2002} are available. In LCSS, the length of the partial overlap is used as the metric. In \cite{Yingchi2017}, the traditional dynamic time warping (DTW) algorithm is used and incorporates several distance factors such as point-segments, consideration of temporal distance factors by converting to spatial distances, and segment-segment distances. However, using only the length of partial overlap as a static threshold for evaluation may lead to non-identification of potential matches \cite{Magdy2017AGT}. All other methods compare two routes completely with each other, which only leads to usable results for scenario [1,1] shown in Figure~\ref{fig:scenarios}, which leads to disadvantages in other valid scenarios. Additional methods of the same type can be found in \cite{Ketabi2018} and \cite{Cruz2015}. Cruz~et~al.~\cite{Cruz2015} cluster discrete points of trajectories, similar to \cite{Chawathe2018}, in order to find a common point on several trajectories. Aydin~et~al.~\cite{Aydin2020} encounter partial route matching using the Needleman-Wunsch-Algorithm \cite{Needleman1970}. Based on distance parameters, in \cite{Schreieck2016} and \cite{Weissig1999}, partial routes are supported, but not the formation of meeting points. Likewise, in \cite{Mallus2017}, meeting points are considered, but only under the condition of pedestrian accessibility by passengers. In \cite{Czioska2017}, additionally, a detour of the driver with a similar travel time as of the passenger walking is included. In \cite{Bitmonnot2013}, \cite{Aissat2015}, and \cite{Haosheng2019}, meeting points with multimodal accessibility are considered, but path similarity between them are not taken into account. Yao and Bekhor~\cite{Yao2021} use a dynamic tree algorithm to match similar routes. 

To the best of our knowledge, the literature provides no algorithm that compares two routes in a parameter-free way, incorporating temporal aspects of it. Accordingly, a new algorithmic development is carried out within the scope of this work. For this purpose, the LCSS approach is developed into a parameter-free, dynamic variant.

\section{Method}
\label{sec:method}

\begin{figure*}[t!]
\centering
\resizebox{\linewidth}{!}{
\includegraphics{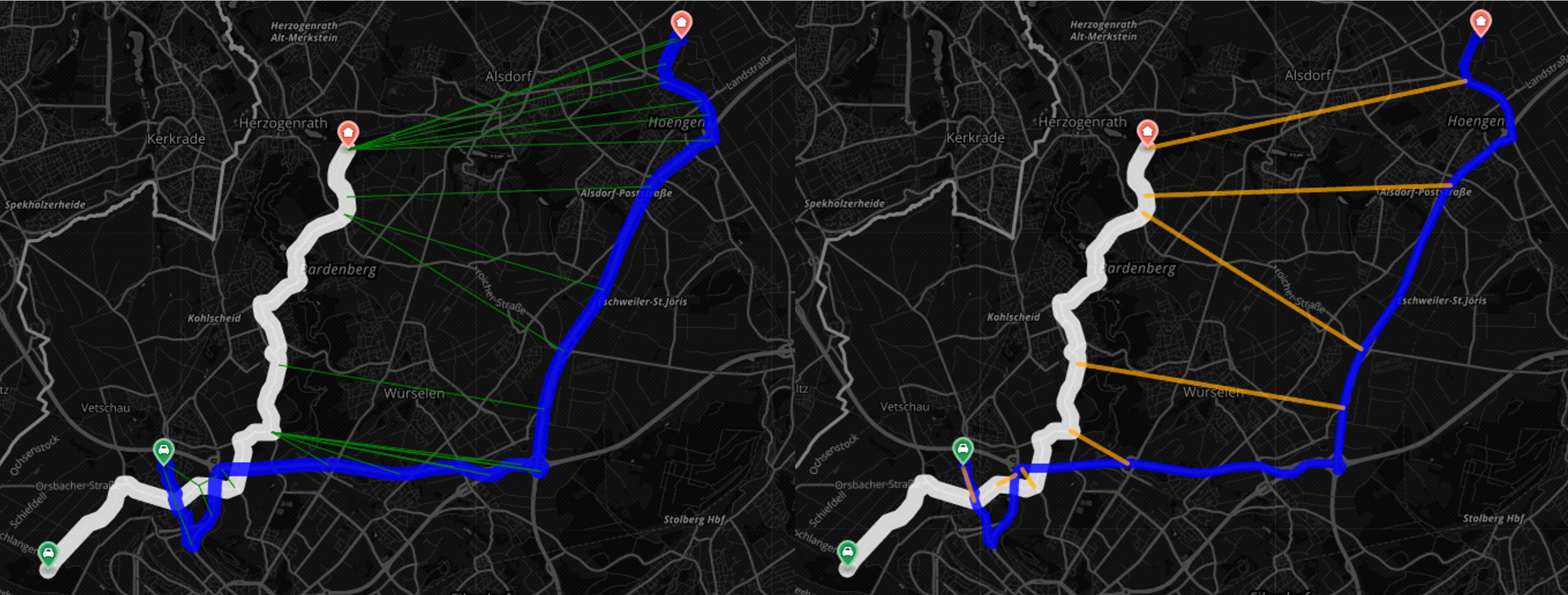}}
\caption{Computation of DLCSS line segments. First, for each coordinate $R_j$ of a requested route (blue), the coordinate $A_i$ of the route of an autonomous vehicle (white) with the shortest distance is identified (left, green lines). Then, for each $A_i$, the shortest line segment is selected, incorporating the temporal route directions (right, orange lines). These DLCSS closed line segments $ls_{DLCSS(A_i,R_j)}$ characterize the similarity of the routes. Adapted from \cite{bogdoll2019MA}.}
\label{fig:dlcss}
\end{figure*}

To check if a route request for an autonomous vehicle is applicable for a shared ride, it needs to be compared with all other planned or existing rides. Depending on the complexity of the underlying matching system, this enables ridesharing with two or more participants. As the evaluation of a match, based on rerouting, is an expensive operation, a lightweight method that provides a similarity metric is desirable. As shown in Figure~\ref{fig:scenarios}, there exist many different scenarios, which need to be taken into account for such a method. Partial overlaps occur particularly often. Generally speaking, it is insufficient to only consider the start and target coordinates for such a comparison, as there may be desired stops along the route or general preferences, e.g., choosing between a highway or a country road.

\begin{algorithm}
\caption{Computation of DLCSS line segments}\label{algo:dlcss}
\KwData{$RouteAutonomous[I]$ and $RouteRequest[J]$ with $I,J \geq 2$ points}
\KwResult{$LineSegmentsDLCSS$ containing 3-tuples $[DistanceDLCSS,p_i,p_j]$}
$LineSegmentsDLCSS[]$\;
$DistanceMatrix[I][J] \gets -1$\;

\ForEach{point $p_j \in RouteRequest$}{
    $MinDistances[]$\;
    \ForEach{point $p_i \in RouteAutonomous$}{
    $MinDistances \gets [distance(p_i,p_j),i]$
    }
    Sort $MinDistances$\;
    $RefI=MinDistances[0][1]$\;
    $DistanceMatrix[RefI][j]=MinDistances[0][0]$
}

$StartJ \gets 0$\;
\ForEach{point $p_i \in RouteAutonomous$}{
    $MinDistances[]$\;
    \For{$j \gets StartJ \in RouteRequest, j \geq RefJ$}{
        \If{$DistanceMatrix[i][j] \neq -1$}{
            $MinDistances \gets [DistanceMatrix[i][j],i,j]$
        }
    }
    Sort $MinDistances$\;
    $StartJ \gets MinDistances[0][2]$\;
    $LineSegmentsDLCSS \gets MinDistances[0]$\;
}
\end{algorithm}

Given a route $A$ from an autonomous vehicle and a requested route $R$ from a potential second passenger, we want to rate the spatio-temporal similarity of the routes based on the partial overlap, but also the distance between these routes within the overlapping section. The proposed DLCSS method, see Algorithm~\ref{algo:dlcss}, allows for a dynamic, yet parameter-free computation of such a score and is visualized in Figure~\ref{fig:dlcss}. In a two-stage minimization approach, first, for each coordinate of $R$, the coordinate of $A$ with the shortest distance is identified. In Algorithm~\ref{algo:dlcss}, this is done in lines $2-11$. At first, however, in line $1$, the result vector is defined. In line $2$, the $DistanceMatrix$ is initialized, which can store the distances between any two points of the routes as an intermediate representation. The following computations can be found in lines $3-11$. Then, for each coordinate reached, the shortest line segment is selected, incorporating the temporal route directions. This is done in lines $12-23$. The $StartJ$ variable from line $12$ is responsible for the temporal component, ensuring that the comparison along $R$ only takes points into account which lie ahead a previously found line segment or the first coordinate, ignoring the past. Then, in lines $13-23$, for each distance stored in the $DistanceMatrix$, the shortest one for each point on $A$ is determined and stored in $LineSegmentsDLCSS$. These DLCSS closed line segments $ls_{DLCSS(A_i,R_j)}$ characterize the similarity of the routes. This way, any combination of two routes can be compared. The algorithm \textbf{dynamically} determines the \textbf{longest common subsequences} for both routes $(A, R)$ and computes a set of closed line segments, which describe the similarity of the routes within these subsequences. Based on this, there are many options available to compute a single metric, quantifying the similarity of two routes. One option we propose is to compute a single similarity metric $sm_{DLCSS}$, which incorporates the percentage overlap of the routes from the point of view of route $A$ as well as the sum of the computed DLCSS closed line segments $ls_{DLCSS(A_i,R_j)}$, see Equation~\ref{eq:dlcss}.

\begin{equation}
\label{eq:dlcss}
    sm_{DLCSS} = \frac{1}{(\frac{l_{Sub_A}}{l_A})} * \sum ls_{DLCSS}
\end{equation}

\begin{figure}[h!]
\centering
\resizebox{\linewidth}{!}{
\includegraphics{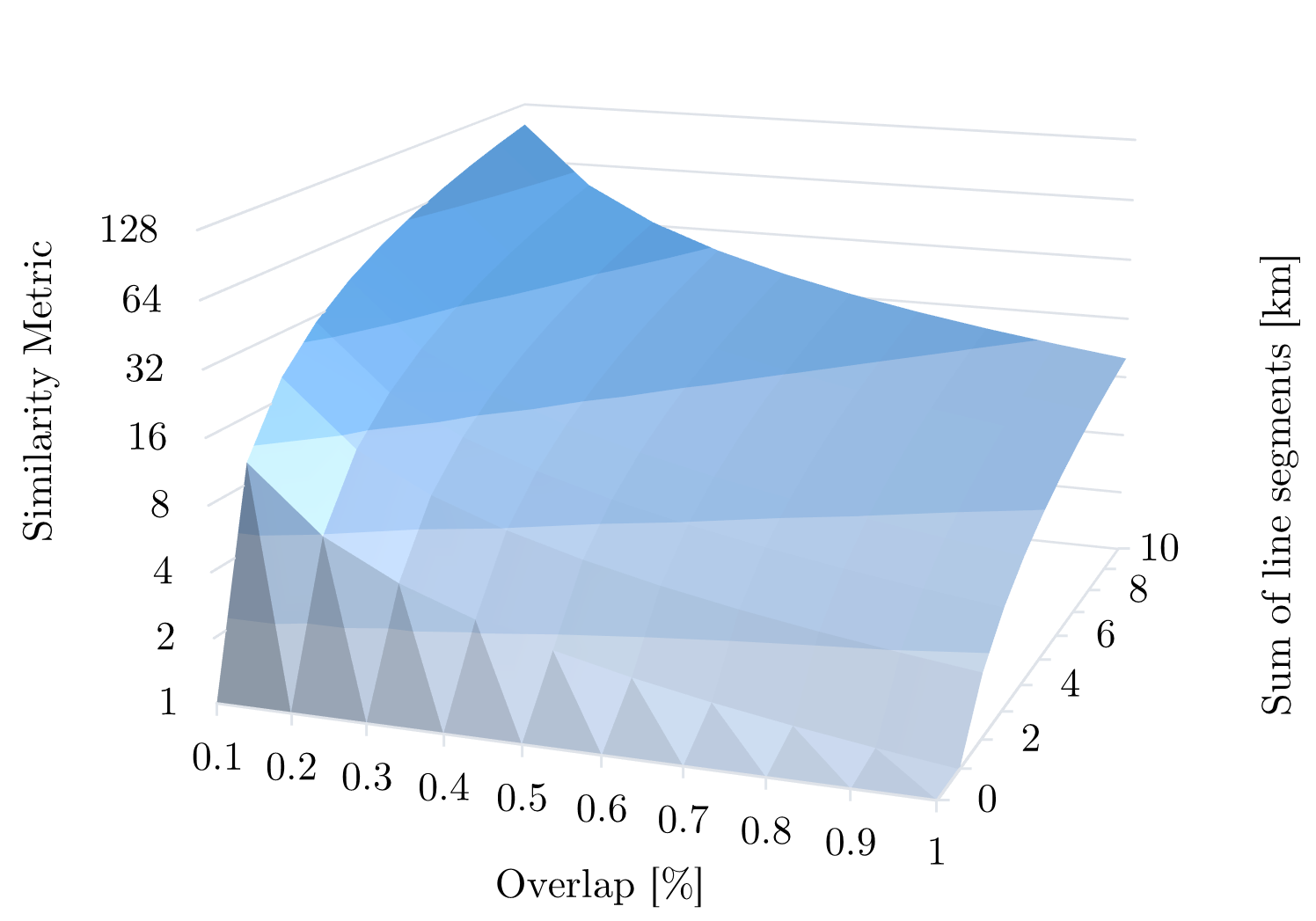}}
\caption{Distribution of $sm_{DLCSS}$ values for a range of line segment sums.}
\label{fig:metric}
\end{figure}

Here, high overlaps and small aggregated line segments lead to a small and thus better similarity metric, while especially high aggregated line segments lead to a high, and thus worse, value. An exemplary distribution is visualized in Figure~\ref{fig:metric}.

\subsection{Application: Meeting Points}

For the five scenarios [*,3] and [3,*] shown in Figure~\ref{fig:scenarios}, a meeting or departure point is necessary for a shared autonomous journey. Computing the potential for a shared ride with a meeting point can be done, when the initial $sm_{DLCSS}$ value for two given routes is too high for a direct shared ride, meaning the detour of the autonomous vehicle would become too large. Based on new routes including a set of meeting points, the $sm_{DLCSS}$ values for each combination can be computed to determine if one of the meeting points might enable a shared ride.

\begin{figure}[h!]
\centering
\resizebox{\linewidth}{!}{
\includegraphics{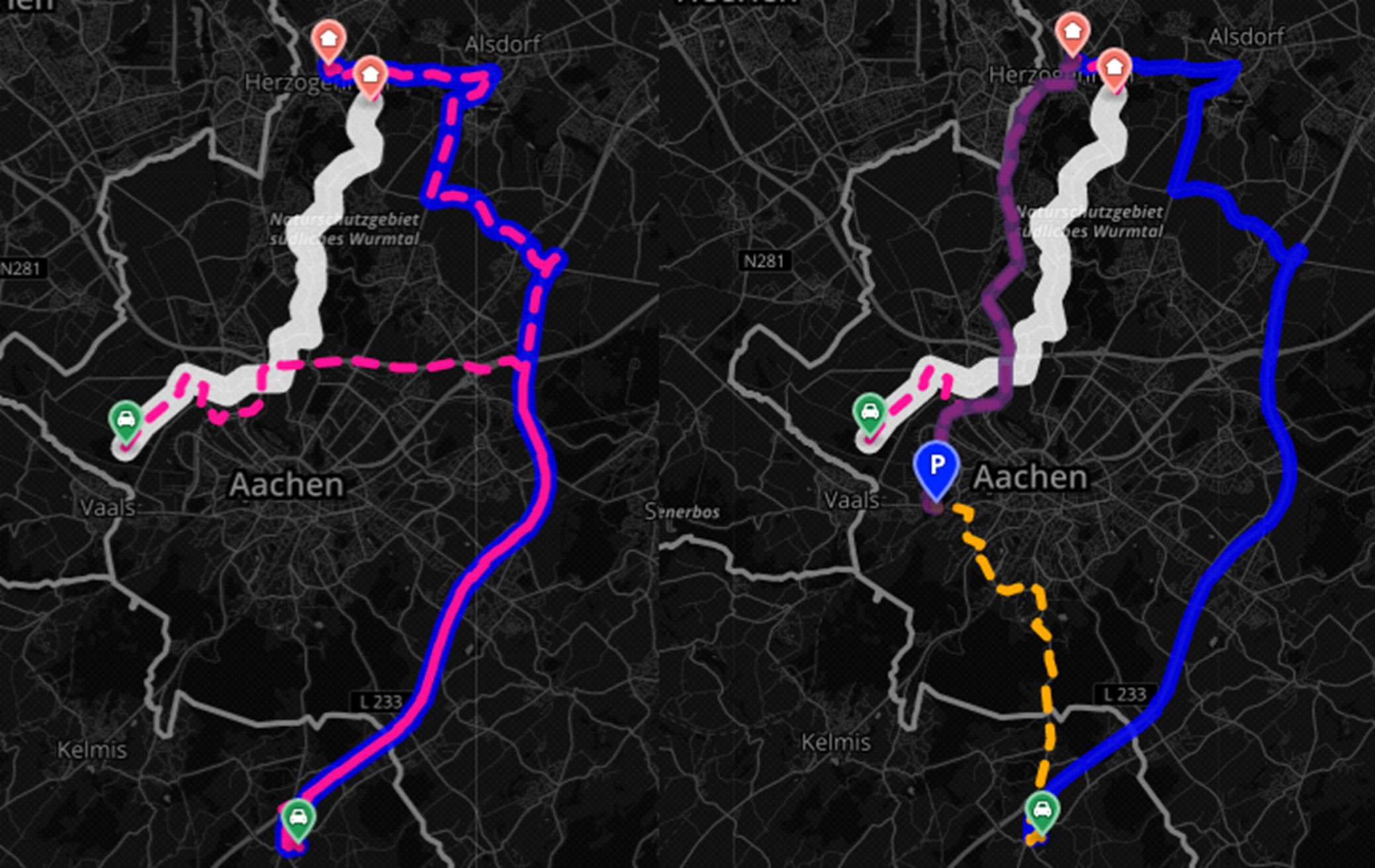}}
\caption{Application of DLCSS to determine if a route can be facilitated with a common meeting point. On the left, the theoretical scenario [1,1] is shown for two routes. On the right, The scenario [1,3] is shown based on a common meeting point. Adapted from \cite{bogdoll2019MA}.}
\label{fig:meeting_point}
\end{figure}

In Figure~\ref{fig:meeting_point}, an exemplary case for this is shown. For every two routes, that show a slightly too large $sm_{DLCSS}$ value for a direct match, the potential of a shared ride with a meeting point can be computed. First, a dataset of potential meeting points was assembled for the Aachen Area in Germany, based on publicly available data \cite{parken1, parken2, parken3}. Here, on the left, a potential shared route with a high $sm_{DLCSS}$ value is shown, which is not suitable. On the right, a meaningful meeting point could be identified.

\section{Evaluation}
\label{sec:evaluation}

\begin{figure}[h!]
\centering
\resizebox{\linewidth}{!}{
\includegraphics{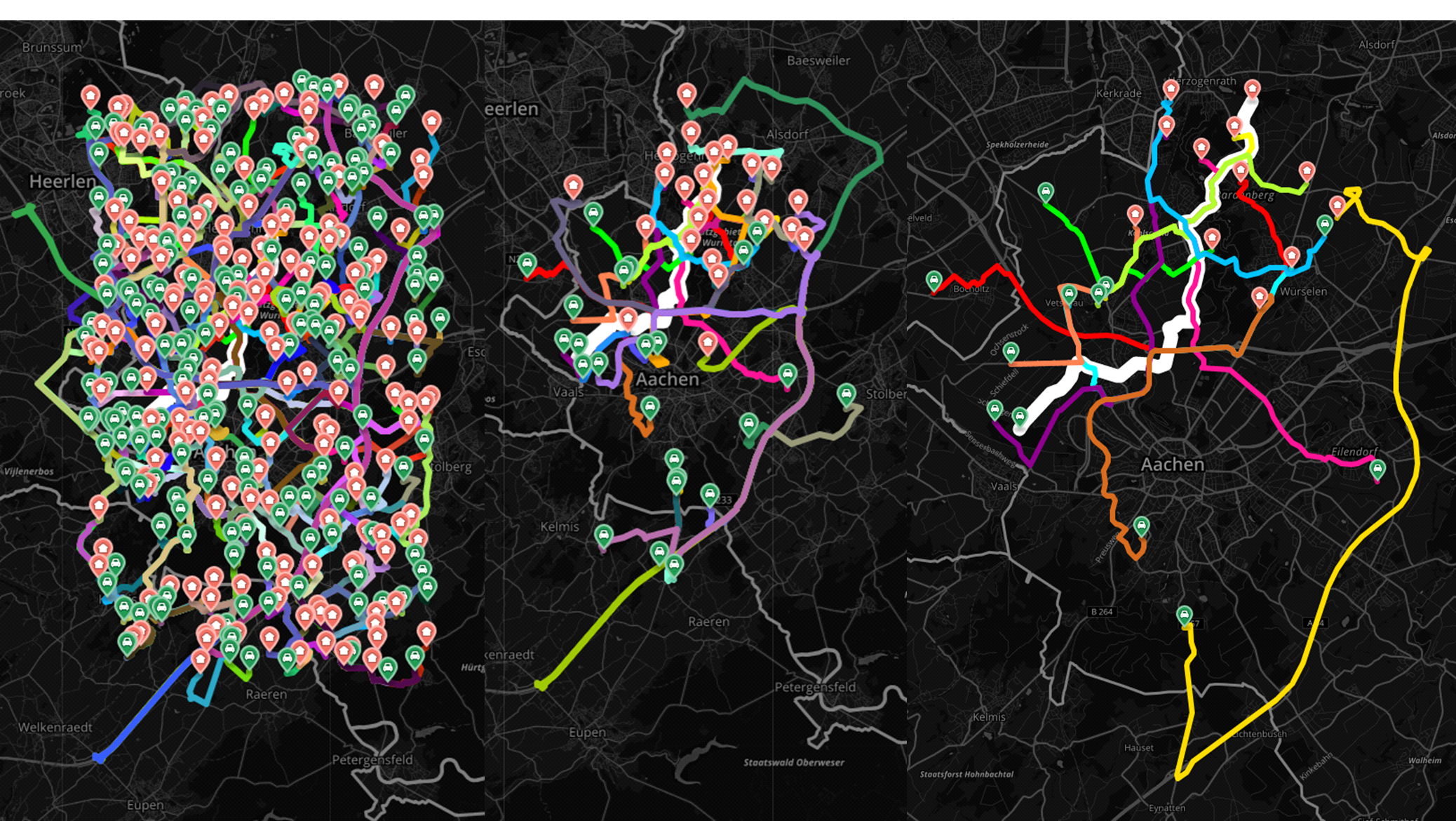}}
\caption{Application of DLCSS to a dataset in the Aachen area~\cite{flate}. The left row shows the whole artificially generated datasets. In the center row, the selection based on the DLCSS algorithm is shown. The right row shows the final result, based on a computationally expensive routing-based matching algorithm. Adapted from \cite{bogdoll2019MA}.}
\label{fig:result}
\end{figure}

To discuss, to which extent the $sm_{DLCSS}$ metric, based on the DLCSS algorithm, is suitable to discard irrelevant route pairs, a quantitative analysis was performed. First, it was necessary to determine a suitable threshold. For that purpose, a dataset with 180 routes in the Aachen area in Germany~\cite{flate} was used. Then, the correlation between the computed $sm_{DLCSS}$ metric and the actual required detour for a shared ride was analyzed, leading to a $sm_{DLCSS}$ value of $20k$ as a suitable threshold, given the assumption, that an autonomous vehicle will only accept detours with up to 50\% the length of its original route. With this, the DLCSS algorithm was able to reject 87.2\% of the entries from the dataset. Here, the comparison with the actual detour showed, that the results contained 47.82\% true positives, while having no false negatives. This overly cautious threshold combines two ideal settings: First, no potential matches are being left out. Second, expensive, routing-based comparisons need only to be made for 12.8\% of the data. Exemplary results are shown in Figure~\ref{fig:result}. For this analysis, no meeting points were included.

\section{Conclusion}
\label{sec:conclusion}

In this work, we provided a literature overview of route matching algorithms for the scenario of shared mobility models for autonomous driving. After proving a holistic scenario matrix for all possible cases, the Dynamic Longest Common Subsequences (DLCSS) algorithm is introduced, which is a dynamic, parameter-free improvement of the LCSS algorithm. In a second step, we introduced the $sm_{DLCSS}$ metric, which is a way of assigning a numeric value to a pair of routes, describing their similarity. Both the proposed algorithm and the metric can be used by operators to improve their fleet operations and  can be a starting point for further analysis. This can help providers to estimate the necessary fleet sizes, policymakers to better understand mobility patterns and cities to scale necessary infrastructure.

In future work, the applicability of the approach for more than two passengers shall be examined. First steps with a greedy merge approach~\cite{bogdoll2019MA} show promising results. Also, the DLCSS raw data can be used to design further metrics, which might lead to improvements in respect to the stability of the similarity metric.
\section{Acknowledgment}
\label{sec:acknowledgment}

This work results partly from the KIGLIS project supported by the German Federal Ministry of Education and Research (BMBF), grant number 16KIS1231.


{\small
\bibliographystyle{IEEEtran}
\bibliography{references}
}

\end{document}